\documentclass[9pt]{extarticle}
\usepackage{spconf,amsmath,amssymb,algorithm2e,algorithmic,graphicx, xcolor, lipsum}
\usepackage{hyperref}
\usepackage[capitalize]{cleveref}
\crefname{section}{Sec.}{Secs.}
\Crefname{section}{Section}{Sections}
\Crefname{table}{Table}{Tables}
\crefname{table}{Tab.}{Tabs.}
\RestyleAlgo{ruled}

\newcommand\blfootnote[1]{%
  \begingroup
  \renewcommand\thefootnote{}\footnote{#1}%
  \addtocounter{footnote}{-1}%
  \endgroup
}

\usepackage{spconf}
\title{Long Range Constraints for Neural Texture Synthesis Using Sliced Wasserstein Loss}
\twoauthors
 {Liping Yin}
	{Michigan State University \\
	CMSE and Department of Mathematics \\
	East Lansing, MI\\
    yinlipi1@msu.edu}
 {Albert Chua}
	{Michigan State University\\
	Department of Mathematics\\
	East Lansing, MI\\
    chuaalbe@msu.edu}

\begin{document}
%
\maketitle
\begin{abstract}
In the past decade, exemplar-based texture synthesis algorithms have seen strong gains in performance by matching statistics of deep convolutional neural networks. However, these algorithms require regularization terms or user-added spatial tags to capture long range constraints in images. Having access to a user-added spatial tag for all situations is not always feasible, and regularization terms can be difficult to tune. Thus, we propose a new set of statistics for texture synthesis based on Sliced Wasserstein Loss, create a multi-scale method to synthesize textures without a user-added spatial tag, study the ability of our proposed method to capture long range constraints, and compare our results to other optimization-based, single texture synthesis algorithms.
\blfootnote{\href{https://github.com/liping1005/LongRangeSlicedWasserstein}{https://github.com/liping1005/LongRangeSlicedWasserstein} for code and textures used in experiments. More examples of synthesis provided as well.}
\blfootnote{We would like to thank Dr. Matthew Hirn for helpful comments and suggestions.}
\blfootnote{This work was supported in part through computational resources and services provided by the Institute for Cyber-Enabled Research at Michigan State University.}
\end{abstract}
\section{Introduction}
\label{sec:intro}
\subsection{Background on exemplar-based texture synthesis}
\label{subsec: background on texture synthesis}
Texture synthesis has been of interest to researchers for over half a century starting with \cite{julesz}. The objective is to take a reference texture and utilize statistical information from the reference texture to generate a different texture with similar perceptual properties as the reference texture. While this problem is inherently interesting as a computer vision task, the literature from this subfield of research also provides us with a better understanding of how humans perceive texture.

Until the emergence of deep convolutional neural networks for image classification, many methods, like \cite{heegerbergen, simoncelli}, have used statistical information of wavelet coefficients for texture generation. However, these approaches failed to produce believable synthesis for images with complex structures, which suggested more statistical information was needed to capture the essence of a texture. 
Neural texture synthesis, which involves the use of deep neural networks to generate textures, started with the seminal work of Gatys et. al. in \cite{stylegatys, texturegatys}. The authors used the mean squared error between gram matrices of VGG19 feature maps as a loss function. While the results in \cite{texturegatys} were state-of-the-art compared to previous work at the time of publication, the proposed method has trouble capturing long range constraints.

Numerous papers have imposed long range constraints to synthesized images for single texture synthesis \cite{spectrum, snelgrove, deepcorr, gonthier, adversarialexpansion}, which generates one texture per synthesis process. However, these methods generally require tunable regularization parameters, do not fully capture long range constraints, or require multiple hours to synthesize one image. Other approaches include universal texture synthesis algorithms \cite{ulyanov, transposedconv, neuralfft}, which train on a database of textures to reduce synthesis time after training; nonetheless, the focus of this paper will be on single texture synthesis.

\subsection{Background on Sliced Wasserstein Loss}
\label{subsec: SW background}
The authors of \cite{SWsynthesis} propose using Sliced Wasserstein Loss. In a manner similar to \cite{heegerbergen, OG_SW}, the authors match the distributions between feature maps, but \cite{SWsynthesis} match distributions between feature maps of VGG19 \cite{VGG19}.

Suppose that layer $\ell$ of an $L$ layer convolutional neural network has $N_\ell$ channels and $M_\ell$ pixels in each channel. We denote the feature vector located at pixel $m$ as $F_{m}^\ell \in \mathbb{R}^{N_\ell}$. With respect a network architecture, let $p^\ell$ and $\hat{p}^\ell$ be the probability density functions associated with the set of feature vectors $\{F_m^\ell\}$ and $\{\hat{F}_m^\ell\}$. We assume that the probability density functions are always an average of Dirac delta distributions of the form
\begin{align}
    p^\ell(x) = \frac{1}{M_\ell} \sum_{m = 1}^{M_\ell} \delta_{F_m^\ell}(x).
\end{align}
Let $V \in \mathbb{S}^{N_\ell}$ be a random direction on the unit sphere of dimension $N_\ell$. The Sliced Wasserstein Distance between two distributions of features is of the form
\begin{align}
\mathcal{L}_{\text{SW}, \ell}(p^\ell, \hat{p}^\ell) = \mathbb{E}_V[\mathcal{L}_{\text{SW1D}}(p^\ell_{V}, \hat{p}^\ell_{V})],
\end{align}
where 
\begin{align} \label{eqn: projections}
p^\ell_V :=\{\langle F_m^{\ell}, V \rangle\}    
\end{align} is a set consisting of batched projections of the feature maps $F_m^{\ell}$ onto $V$; define vector $P_V^\ell$ consisting of the elements of $p^\ell_V$, the $1D$ Sliced Wasserstein Loss is the $2$-norm between sets of sorted projections:
\begin{align} 
\mathcal{L}_{\text{SW1D}}(p^\ell_{V}, \hat{p}^\ell_{V}) = \frac{1}{\text{len}(P_V^\ell)}\left\| \text{sort}(P_V^\ell) -\text{sort}(\hat{P}_V^\ell)\right\|_2^2
\end{align}
and the full Sliced Wasserstein loss over all the layers is
\begin{align} \label{eqn: sliced wasserstein channels}
\mathcal{L}_{\text{SW}}(I_1, I_2) = \sum_{\ell=1}^{L}w_i \mathcal{L}_{\text{SW},\ell}(p^\ell_{V, I_1}, p^\ell_{V, I_2}),
\end{align}
for images $I_1$ and $I_2$, respectively, where $w_i$ are weight terms that set to zero for layers that are not used. 

\subsection{Theoretical justifications for SW loss}
Pitie et al. \cite{pitie2005n} showed that Sliced Wasserstein Distance satisfies:$$\mathcal{L}_{\text{SW}}(p, \hat{p}) = 0 \implies p = \hat{p}.$$
The same does not hold for other losses used for texture synthesis, such as the gram matrix loss. Thus, using a Sliced Wasserstein-based loss should capture  more stationary statistics compared to the traditional Gram Loss. 

\subsection{Our contributions}
While theoretically justified, the method proposed in \cite{SWsynthesis} cannot effectively capture long range constraints unless a user-added spatial tag is added to guide synthesis. We propose a new set of statistics for single texture synthesis based on Sliced Wasserstein Loss to capture long range constraints without any supervision or hyperparameter tuning. The proposed set of statistics displays competitive results compared to other single texture synthesis algorithms, and we augment our synthesis results via a coarse-to-fine multi-scale procedure. As a short aside, the primary purpose of this paper is not to get state-of-the art synthesis results; our goal is to find a minimal set of unsupervised statistics that capture texture in an image. 

\section{Texture Synthesis Algorithm}
Instead of matching distributions via slicing over the channel dimension of the feature maps, we purpose adding another set of statistics to be matched. Consider a set of feature maps $F^\ell \in \mathbb{R}^{H_\ell \times W_\ell \times N_\ell}$. In \cite{SWsynthesis}, one unravels each $H_\ell\times W_\ell$ feature map, and projects the feature vector associated to each pixel onto direction $V$ in \cref{eqn: sliced wasserstein channels}. 

To add another loss term, reshape the feature maps into $H_\ell$ different $W_\ell \times N_\ell$ feature vectors, $F_{H}^\ell$ (with $F_{H,n}^\ell$ being a vector of all $n^{\text{th}}$ pixels of each feature vector), and project them onto $V_{H_\ell} \in \mathbb{S}^{H_\ell}$. Analogous to \cref{eqn: projections}, for the distribution $p_{H}^{\ell}$ associated to feature vectors $\{F_{H,n}^\ell\}$, define
\begin{align}
p_{V_{H_\ell}}^\ell =  \{\langle F_{H,n}^\ell, V_{H_\ell} \rangle\}. 
\end{align}
The corresponding additional loss term is 
\begin{align} \label{eqn: sliced wasserstein height}
\mathcal{L}_{\text{SW}, H}(I_1, I_2) = \sum_{\ell=1}^{L} w_i\mathcal{L}_{\text{SW},\ell}\left(p^\ell_{V_{H_\ell}, I_1}, p^\ell_{V_{H_\ell}, I_2}\right).
\end{align}
Our new loss function is
\begin{align} \label{eqn: new slicing loss}
 \mathcal{L}_\text{Slicing}(I_1,I_2) =  \mathcal{L}_{\text{SW}}(I_1, I_2) + \mathcal{L}_{\text{SW}, H}(I_1, I_2),
\end{align} which is the sum of \cref{eqn: sliced wasserstein channels} and  \cref{eqn: sliced wasserstein height}. 

Denote the feature map extraction from VGG19 as $\text{Extract}(I)$. Start with a reference image $I_{\text{ref}}$ and a white noise $I_{\text{WN}}$ and run for $M$ epochs. The implementation for slicing synthesis is the same as in \cite{SWsynthesis} for \cref{eqn: sliced wasserstein channels}. For the additional loss term in equation \cref{eqn: sliced wasserstein height}, the number of batched projections is $H_\ell$. 

In the next algorithm, assume that the goal is to synthesize an image the same size as the reference image without any loss of generality. The settings for the slicing loss are to use the first 12 layers of VGG19 for calculating $\mathcal{L}_{\text{SW}}$ and the first two convolutions in each convolution block for for calculating $\mathcal{L}_{\text{SW},H}$. The L-BFGS optimizer is used for optimization with a learning rate of $\eta = 1$. 
\begin{algorithm}
\caption{Synthesis Algorithm}\label{alg:one}
\begin{algorithmic}[1]
\STATE Set $I_{\text{WN}}$ as variable to be updated by optimizer. 
\FOR{$k = 1, \ldots, M$}
    \STATE Calculate $\text{Extract}(I_{\text{WN}})$.
    \STATE Calculate $\text{Extract}(I_{\text{ref}})$. 
    \STATE Calculate $\mathcal{L}_\text{Slicing}(I_{\text{WN}},I_{\text{ref}})$. 
    \STATE Backpropogate and update $I_{\text{WN}}$.
\ENDFOR
\STATE Return updated $I_{\text{WN}}$ as synthesized texture.
\end{algorithmic}
\end{algorithm}

\subsection{Comparison with Other Methods}
To test our proposed algorithm, we compare our results with algorithms that have similar runtime and computational complexity. A comparison with Heitz. et. al.\footnote{We use the author's TensorFlow implementation, which is a previous commit in https://github.com/tchambon/A-Sliced-Wasserstein-Loss-for-Neural-Texture-Synthesis.} (without a spatial tag) and with gram matrices using a spectrum constraint for generating simple, psuedoperiodic $256 \times 256$ textures\footnote{For all the experiments in this paper, our texture sources were the following: \href{https://github.com/omrysendik/DCor/tree/master/Data}{the DeepCorr Github Page} and \href{https://www.robots.ox.ac.uk/~vgg/data/dtd}{the DTD texture database.}} with long range constraints are given in Fig. 1. 

\begin{figure}[hbt]
\begin{minipage}[t]{0.2375\columnwidth}
  \includegraphics[width=\linewidth]{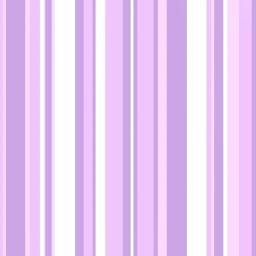}
\end{minipage}\hfill 
\begin{minipage}[t]{0.2375\columnwidth}
\includegraphics[width=\linewidth]{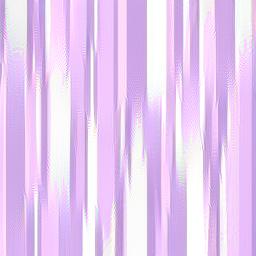}
\end{minipage}\hfill 
\begin{minipage}[t]{0.2375\columnwidth}
\includegraphics[width=\linewidth]{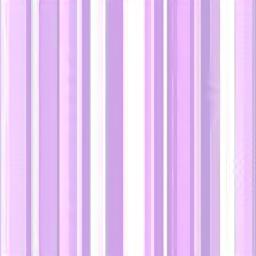}
\end{minipage}\hfill 
\begin{minipage}[t]{0.2375\columnwidth}
  \includegraphics[width=\linewidth]{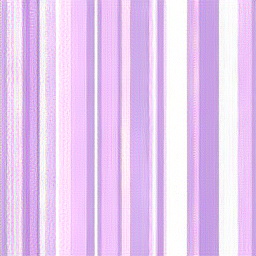}
\end{minipage}
\begin{minipage}[t]{0.2375\columnwidth}
  \includegraphics[width=\linewidth]{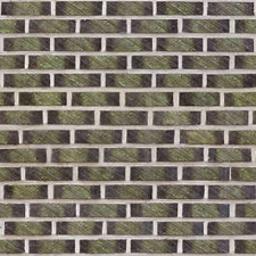}
\end{minipage}\hfill 
\begin{minipage}[t]{0.2375\columnwidth}
\includegraphics[width=\linewidth]{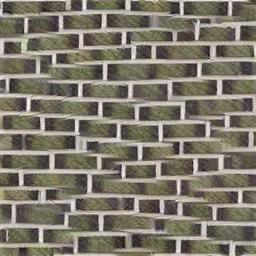}
\end{minipage}\hfill 
\begin{minipage}[t]{0.2375\columnwidth}
  \includegraphics[width=\linewidth]{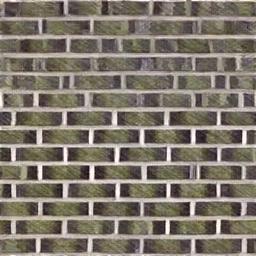}
\end{minipage}\hfill 
\begin{minipage}[t]{0.2375\columnwidth}
  \includegraphics[width=\linewidth]{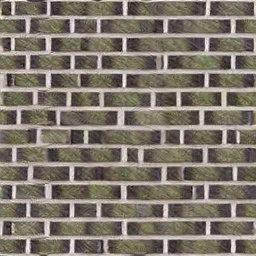}
\end{minipage}\hfill 
\begin{minipage}[t]{0.2375\columnwidth}
  \includegraphics[width=\linewidth]{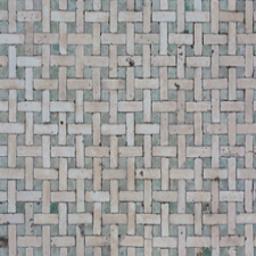}
\end{minipage}\hfill 
\begin{minipage}[t]{0.2375\columnwidth}
\includegraphics[width=\linewidth]{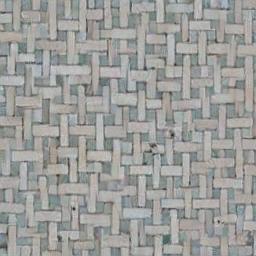}
\end{minipage}\hfill 
\begin{minipage}[t]{0.2375\columnwidth}
  \includegraphics[width=\linewidth]{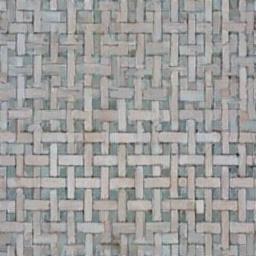}
\end{minipage}\hfill 
\begin{minipage}[t]{0.2375\columnwidth}
  \includegraphics[width=\linewidth]{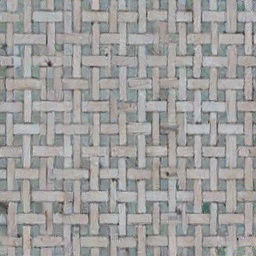}
\end{minipage}\hfill 
\begin{minipage}[t]{0.2375\columnwidth}
  \includegraphics[width=\linewidth]{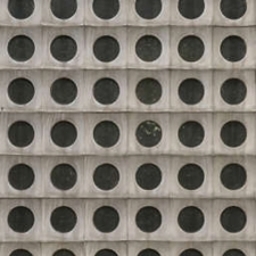}
\end{minipage}\hfill 
\begin{minipage}[t]{0.2375\columnwidth}
\includegraphics[width=\linewidth]{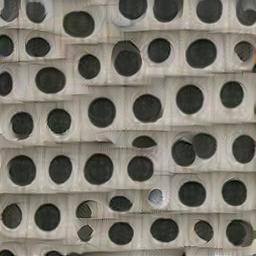}
\end{minipage}\hfill 
\begin{minipage}[t]{0.2375\columnwidth}
  \includegraphics[width=\linewidth]{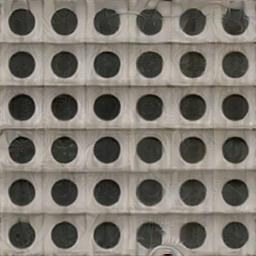}
\end{minipage}\hfill 
\begin{minipage}[t]{0.2375\columnwidth}
  \includegraphics[width=\linewidth]{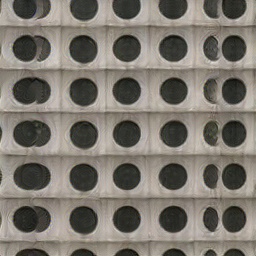}
\end{minipage}\hfill 
\begin{minipage}[t]{0.2375\columnwidth}
  \includegraphics[width=\linewidth]{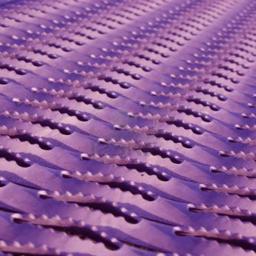}
\end{minipage}\hfill 
\begin{minipage}[t]{0.2375\columnwidth}
\includegraphics[width=\linewidth]{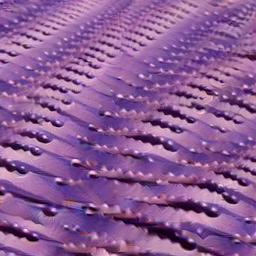}
\end{minipage}\hfill 
\begin{minipage}[t]{0.2375\columnwidth}
  \includegraphics[width=\linewidth]{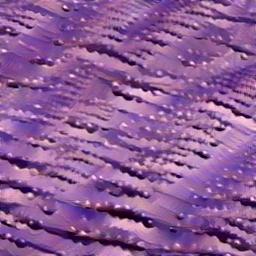}
\end{minipage}\hfill 
\begin{minipage}[t]{0.2375\columnwidth}
  \includegraphics[width=\linewidth]{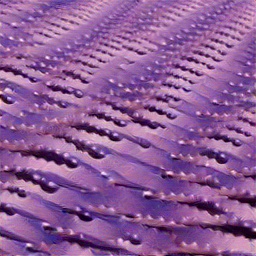}
\end{minipage}\hfill 
\caption{Comparison of results for pseudoperiodic textures. \textbf{Left:} Reference. \textbf{Mid Left:} SW Loss. \textbf{Mid Right:} Spectrum. \textbf{Right:} Using \cref{eqn: new slicing loss} [Ours].}
\label{fig: comparison K = 0, stationary} 
\end{figure} 
From the results in Fig. 1, using \cref{eqn: new slicing loss} yields long range constraints, and the results are comparable to \cite{spectrum}. However, note that the synthesis using \cref{eqn: new slicing loss} yields less faithful synthesis relative to \cite{spectrum} in some cases, such as in the fourth row, but can also yield better synthesis compared to \cite{spectrum} in other cases, such as in the fifth row. One possible reason for the failure of using \cref{eqn: new slicing loss} is using \cref{eqn: sliced wasserstein height} captures only horizontal stationary statistics for each tensor of VGG19 feature maps, but not vertical stationary statistics. 

However, many textures have nonstationary components, so a comparison with more complex textures is in Fig 2. 
\begin{figure}[h]
\begin{minipage}[t]{0.2375\columnwidth}
  \includegraphics[width=\linewidth]{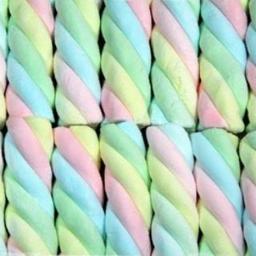}
\end{minipage}\hfill 
\begin{minipage}[t]{0.2375\columnwidth}
\includegraphics[width=\linewidth]{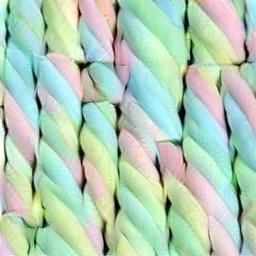}
\end{minipage}\hfill 
\begin{minipage}[t]{0.2375\columnwidth}
\includegraphics[width=\linewidth]{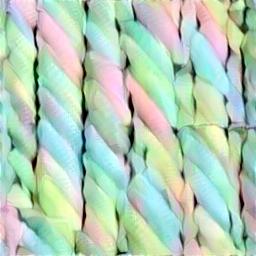}
\end{minipage}\hfill 
\begin{minipage}[t]{0.2375\columnwidth}
  \includegraphics[width=\linewidth]{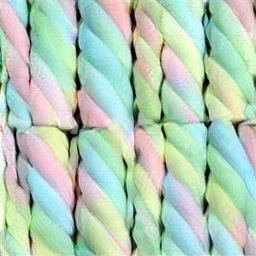}
\end{minipage}
\begin{minipage}[t]{0.2375\columnwidth}
  \includegraphics[width=\linewidth]{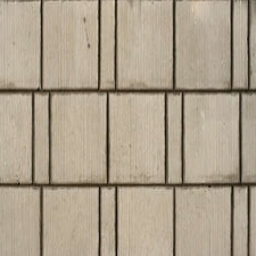}
\end{minipage}\hfill 
\begin{minipage}[t]{0.2375\columnwidth}
\includegraphics[width=\linewidth]{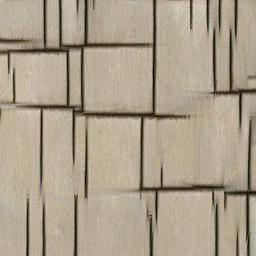}
\end{minipage}\hfill 
\begin{minipage}[t]{0.2375\columnwidth}
  \includegraphics[width=\linewidth]{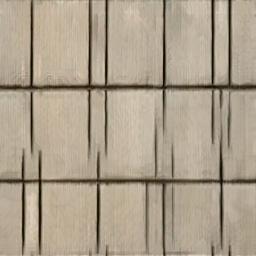}
\end{minipage}\hfill 
\begin{minipage}[t]{0.2375\columnwidth}
  \includegraphics[width=\linewidth]{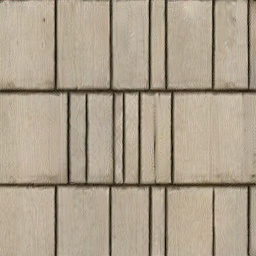}
\end{minipage}\hfill 
\begin{minipage}[t]{0.2375\columnwidth}
  \includegraphics[width=\linewidth]{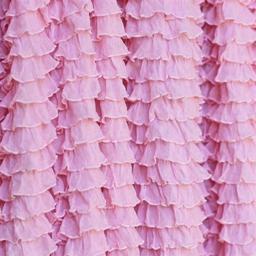}
\end{minipage}\hfill 
\begin{minipage}[t]{0.2375\columnwidth}
\includegraphics[width=\linewidth]{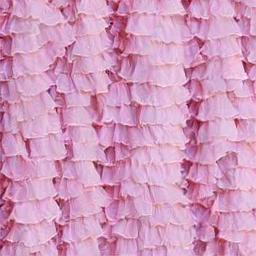}
\end{minipage}\hfill 
\begin{minipage}[t]{0.2375\columnwidth}
  \includegraphics[width=\linewidth]{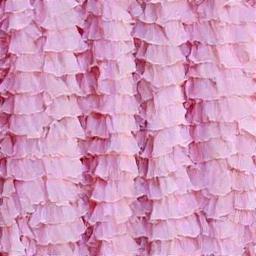}
\end{minipage}\hfill 
\begin{minipage}[t]{0.2375\columnwidth}
  \includegraphics[width=\linewidth]{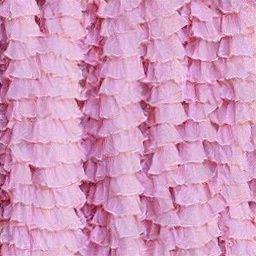}
\end{minipage}\hfill 
\begin{minipage}[t]{0.2375\columnwidth}
  \includegraphics[width=\linewidth]{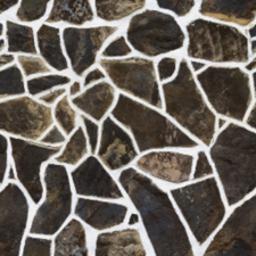}
\end{minipage}\hfill 
\begin{minipage}[t]{0.2375\columnwidth}
\includegraphics[width=\linewidth]{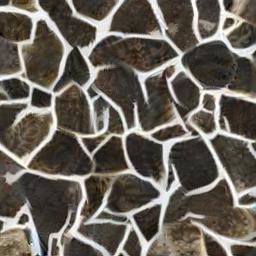}
\end{minipage}\hfill 
\begin{minipage}[t]{0.2375\columnwidth}
  \includegraphics[width=\linewidth]{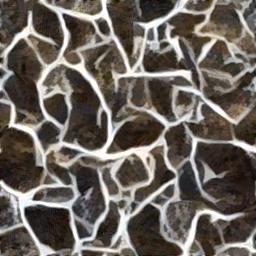}
\end{minipage}\hfill 
\begin{minipage}[t]{0.2375\columnwidth}
  \includegraphics[width=\linewidth]{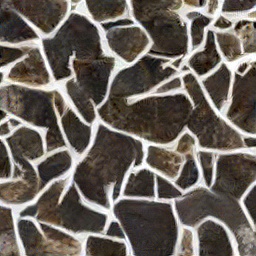}
\end{minipage}\hfill 
\begin{minipage}[t]{0.2375\columnwidth}
  \includegraphics[width=\linewidth]{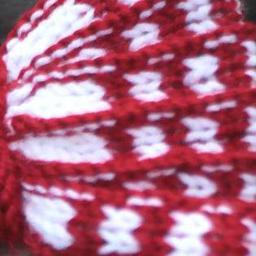}
\end{minipage}\hfill 
\begin{minipage}[t]{0.2375\columnwidth}
\includegraphics[width=\linewidth]{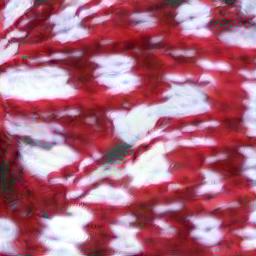}
\end{minipage}\hfill 
\begin{minipage}[t]{0.2375\columnwidth}
  \includegraphics[width=\linewidth]{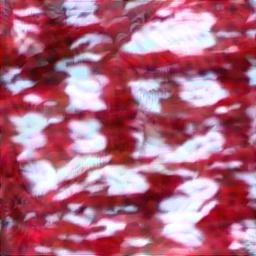}
\end{minipage}\hfill 
\begin{minipage}[t]{0.2375\columnwidth}
  \includegraphics[width=\linewidth]{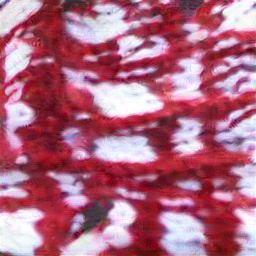}
\end{minipage}\hfill 
\begin{minipage}[t]{0.2375\columnwidth}
  \includegraphics[width=\linewidth]{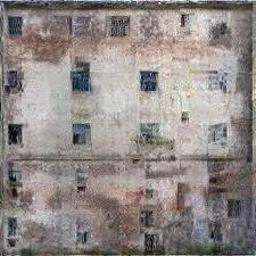}
\end{minipage}\hfill 
\begin{minipage}[t]{0.2375\columnwidth}
\includegraphics[width=\linewidth]{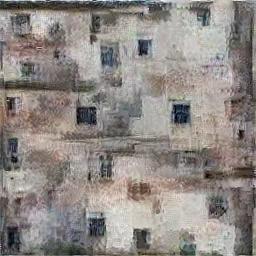}
\end{minipage}\hfill 
\begin{minipage}[t]{0.2375\columnwidth}
  \includegraphics[width=\linewidth]{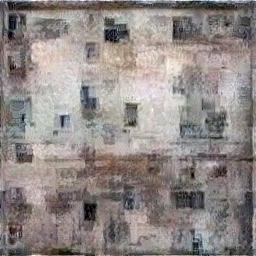}
\end{minipage}\hfill 
\begin{minipage}[t]{0.2375\columnwidth}
  \includegraphics[width=\linewidth]{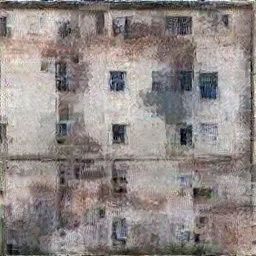}
\end{minipage}\hfill 
\caption{Comparison of results for more complex textures. \textbf{Left:} Reference. \textbf{Mid Left:} SW Loss. \textbf{Mid Right:} Spectrum. \textbf{Right:} Using \cref{eqn: new slicing loss} [Ours].}
\label{fig: comparison K = 0, nonstationary} 
\end{figure}
From the results of Fig. 2, it is apparent that our proposed algorithm yields better synthesis results on textures with nonstationary components. 

To quantify our results, a set of 34 textures were compiled for a quantitative study between the original SW loss, the spectrum constraint, and our proposed method in Table \ref{Tab: quant comparison}. Most of the textures in this set have long range constraints, some pseudoperiodic and some with nonstationary components. The textures from this set are also used throughout for the visual comparisons in the paper. In the table, SW stands for the method using the original SW Loss, Spec. stands for using a spectrum constraint, and GT stands for the Ground Truth. 

We chose the following quantitative metrics: LPIPS \cite{LPIPS}, FID \cite{FID}, crop-based FID (this done by taking sixty-four $128 \times 128$ crops of the reference texture and synthesized texture for each exemplar. The FID score is calculated between these two sets of images. For the ground truth case, a different set of crops of the reference is used) similar to \cite{transposedconv}, KID \cite{KID}, and crop-based KID (c-KID) score. For FID and KID based scores, the implementation from \cite{cleanfid} is used.

\begin{table}[h]
\caption{Quantitative Comparison by Algorithm}
\centering
\begin{tabular}{||c c c c c c||} 
 \hline
 Method & LPIPS & FID & c-FID & KID & c-KID \\ [0.25ex] 
 \hline\hline
 Ours & $\mathbf{0.437}$ & $107.220$ & $\mathbf{71.938}$ & $-0.014$ & $\mathbf{0.073}$\\ 
 \hline
 SW  & $0.454$ & $101.768$ & $78.683$ & $\mathbf{-0.016}$ & $0.083$\\
 \hline
 Spec. & $0.447$ & $\mathbf{99.615}$ & $78.250$ & $\mathbf{-0.016}$& $0.083$\\
 \hline\hline
 GT & $0$ & $0$ & $18.069$ & $-0.025$& $0$\\
 \hline
\end{tabular}
\label{Tab: quant comparison}
\end{table}
From the table, our results are competitive and our proposed set of statistics \textit{did not require searching for a proper hyperparameter to get competitive results}. Note that our results for FID and c-FID cannot be compared to other works because FID is a biased estimator \cite{chong2020} and our sample count is lower.

\subsection{Ablation Study: Varying the Number of Slices}
For the additional loss term in equation \cref{eqn: sliced wasserstein height}, we demonstrate the effect of varying the number of batched projections. Rather than varying the number of slices in each layer by using $H_\ell$ projections, we use $16$, $64$, and $256$ projections. The qualitative results are presented in Fig. \ref{fig: albation, slice count}.
\begin{figure}[h]
\begin{minipage}[t]{0.2375\columnwidth}
  \includegraphics[width=\linewidth]{refs/img_9.jpg}
\end{minipage}\hfill
\begin{minipage}[t]{0.2375\columnwidth}
  \includegraphics[width=\linewidth]{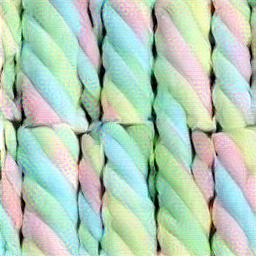}
\end{minipage}\hfill 
\begin{minipage}[t]{0.2375\columnwidth}
\includegraphics[width=\linewidth]{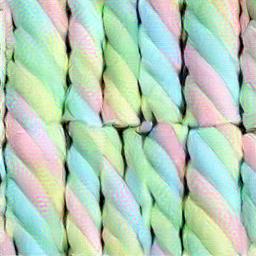}
\end{minipage}\hfill 
\begin{minipage}[t]{0.2375\columnwidth}
\includegraphics[width=\linewidth]{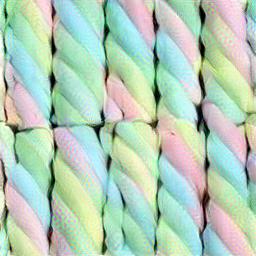}
\end{minipage}
\begin{minipage}[t]{0.2375\columnwidth}
  \includegraphics[width=\linewidth]{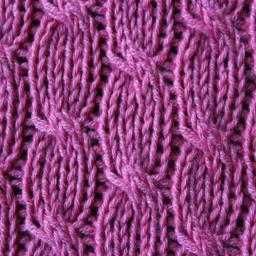}
\end{minipage}\hfill
\begin{minipage}[t]{0.2375\columnwidth}
  \includegraphics[width=\linewidth]{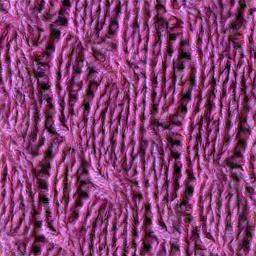}
\end{minipage}\hfill 
\begin{minipage}[t]{0.2375\columnwidth}
\includegraphics[width=\linewidth]{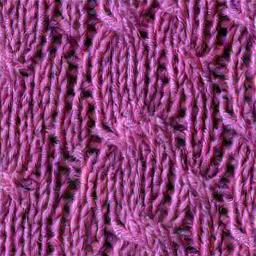}
\end{minipage}\hfill 
\begin{minipage}[t]{0.2375\columnwidth}
\includegraphics[width=\linewidth]{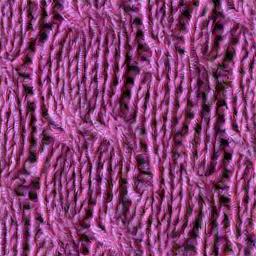}
\end{minipage}
\begin{minipage}[t]{0.2375\columnwidth}
  \includegraphics[width=\linewidth]{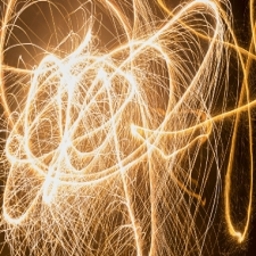}
\end{minipage}\hfill
\begin{minipage}[t]{0.2375\columnwidth}
  \includegraphics[width=\linewidth]{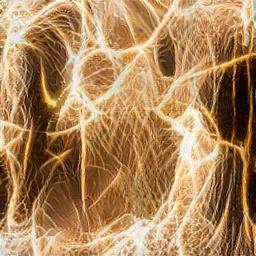}
\end{minipage}\hfill 
\begin{minipage}[t]{0.2375\columnwidth}
\includegraphics[width=\linewidth]{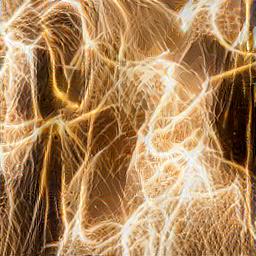}
\end{minipage}\hfill 
\begin{minipage}[t]{0.2375\columnwidth}
\includegraphics[width=\linewidth]{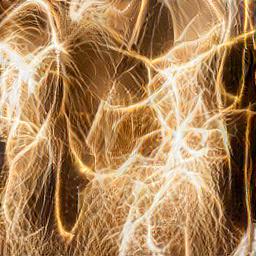}
\end{minipage}
\begin{minipage}[t]{0.2375\columnwidth}
  \includegraphics[width=\linewidth]{refs/img_36.jpg}
\end{minipage}\hfill
\begin{minipage}[t]{0.2375\columnwidth}
  \includegraphics[width=\linewidth]{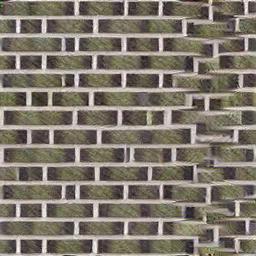}
\end{minipage}\hfill 
\begin{minipage}[t]{0.2375\columnwidth}
\includegraphics[width=\linewidth]{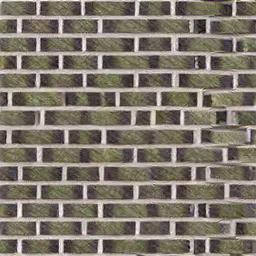}
\end{minipage}\hfill 
\begin{minipage}[t]{0.2375\columnwidth}
\includegraphics[width=\linewidth]{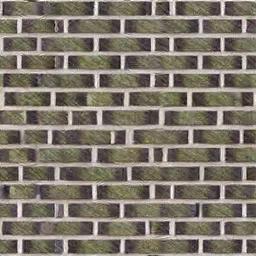}
\end{minipage}
\caption{Textures Generated using a different number of directions. \textbf{Left:} Reference. \textbf{Mid Left:} $16$ directions. \textbf{Mid Right:} $64$ directions. \textbf{Right:} $256$ directions.}
\label{fig: albation, slice count} 
\end{figure}
We also present a table of the runtime as a function of the number of slices in Table \ref{Tab: quant comparison slices}. A total of five runs were done for each image.
\begin{table}[hbt]
\caption{Runtime by Number of Slices (seconds)}
\centering
\begin{tabular}{||c c c c c||} 
 \hline
 Slices & Top & Mid-Top & Mid Bottom & Bottom \\ [0.25ex] 
 \hline\hline
 $16$  &  $37.5 \pm 4.0$ &  $33.0 \pm 4.0$ &  $31.2 \pm 0.6$ &  $30.8 \pm 0.4$\\ 
 \hline
 $64$   & $33.2 \pm 1.0$ &  $31.4 \pm 0.8$&  $31.11 \pm 1.2$ & $32.1 \pm 1.4$\\
 \hline
 $256$  & $45.5 \pm 1.6$ & $46.9 \pm 1.2$ & $46.7 \pm 0.9$ &  $45.6 \pm 1.1$\\
 \hline
$H_\ell$ & $36.8 \pm 4.9$ & $34.3 \pm 1.9$ & $33.8 \pm 1.3$ & $33.3 \pm 1.0$ \\
 \hline\hline
None  &  $27.4 \pm 4.3$ &  $ 28.1 \pm 4.3$ & $25.7 \pm 0.9$ & $28.7 \pm 5.4$ \\
 \hline
\end{tabular}
\label{Tab: quant comparison slices}
\end{table} We can see that our additional loss term yields a small increase in runtime for a marked increase in visual quality, and our choice for the number of random directions in each layer yields similar quality compared to using more random directions.

\section{Improvements via a Multi-scale Approach}
 To ameliorate deficiencies in synthesis quality, we propose augmenting our synthesis process with a multi-scale procedure in a manner similar to \cite{gonthier, kwatra, OG_SW}. The pseudocode is given in Algorithm \ref{alg:two}.
\begin{algorithm}[hbt]
\caption{Multi-scale Synthesis Algorithm}\label{alg:two}
\begin{algorithmic}[1]
\STATE Initialize $I_{\text{Synthesis}}$ as a white noise as ref. texture downsampled by $2^K$ and $I_{\text{ref},i}$ be the ref. texture downsampled by $2^{i}$.
\FOR{$i = 0, \ldots, K$}
    \STATE $I_{\text{Synthesis}} \leftarrow \text{SWSynthesis}(I_{\text{Synthesis}}, I_{\text{ref},{K-i}})$.
    \STATE $I_{\text{Synthesis}} \leftarrow 2\times\text{ Upsample}(I_{\text{Synthesis}})$.
\ENDFOR
\STATE Return $I_{\text{Synthesis}}$ as the synthesized texture.
\end{algorithmic}
\end{algorithm} 

\begin{figure} [bht]
\begin{minipage}[t]{0.2375\columnwidth}
  \includegraphics[width=\linewidth]{refs/img_22.jpg}
\end{minipage}\hfill 
\begin{minipage}[t]{0.2375\columnwidth}
  \includegraphics[width=\linewidth]{k_eq_0/result_22.jpg}
\end{minipage}\hfill
\begin{minipage}[t]{0.2375\columnwidth}
  \includegraphics[width=\linewidth]{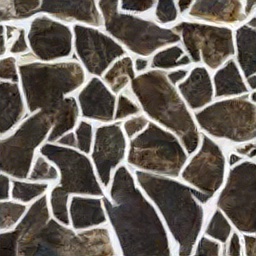}
\end{minipage}\hfill 
\begin{minipage}[t]{0.2375\columnwidth}
  \includegraphics[width=\linewidth]{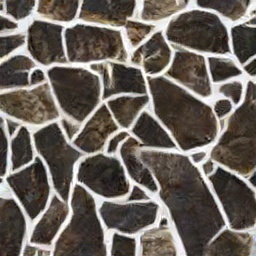}
\end{minipage}\hfill 
\begin{minipage}[t]{0.2375\columnwidth}
  \includegraphics[width=\linewidth]{refs/img_23.jpg}
\end{minipage}\hfill 
\begin{minipage}[t]{0.2375\columnwidth}
  \includegraphics[width=\linewidth]{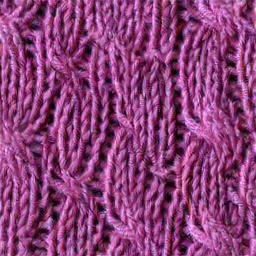}
\end{minipage}\hfill
\begin{minipage}[t]{0.2375\columnwidth}
  \includegraphics[width=\linewidth]{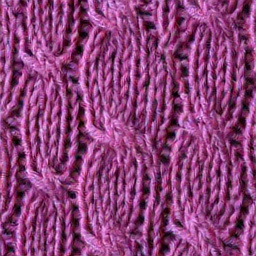}
\end{minipage}\hfill 
\begin{minipage}[t]{0.2375\columnwidth}
  \includegraphics[width=\linewidth]{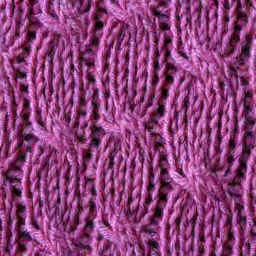}
\end{minipage}\hfill 
\begin{minipage}[t]{0.2375\columnwidth}
  \includegraphics[width=\linewidth]{refs/img_18.jpg}
\end{minipage}\hfill 
\begin{minipage}[t]{0.2375\columnwidth}
  \includegraphics[width=\linewidth]{k_eq_0/result_18.jpg}
\end{minipage}\hfill
\begin{minipage}[t]{0.2375\columnwidth}
  \includegraphics[width=\linewidth]{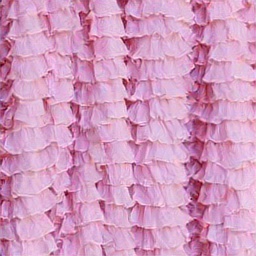}
\end{minipage}\hfill 
\begin{minipage}[t]{0.2375\columnwidth}
  \includegraphics[width=\linewidth]{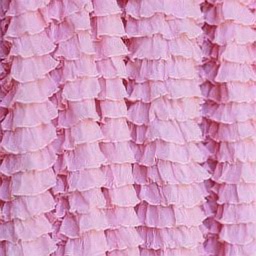}
\end{minipage}
\begin{minipage}[t]{0.2375\columnwidth}
  \includegraphics[width=\linewidth]{refs/img_31.jpg}
\end{minipage}\hfill 
\begin{minipage}[t]{0.2375\columnwidth}
  \includegraphics[width=\linewidth]{k_eq_0/result_31.jpg}
\end{minipage}\hfill
\begin{minipage}[t]{0.2375\columnwidth}
  \includegraphics[width=\linewidth]{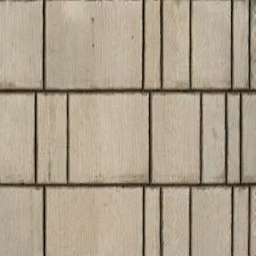}
\end{minipage}\hfill 
\begin{minipage}[t]{0.2375\columnwidth}
  \includegraphics[width=\linewidth]{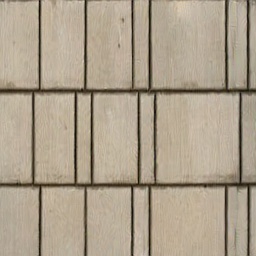}
\end{minipage}\hfill 
\caption{Multi-scale procedure at different scales. \textbf{Left:} Reference. \textbf{Mid Left:} $K = 0$. \textbf{Mid Right:} $K = 1$. \textbf{Right:} $K = 2$.}
\label{fig: mutliscale examples}
\end{figure} 
In Fig. \ref{fig: mutliscale examples}, note the small improvements in edge generation and general structure when using $K = 1$ compared to $K = 0$. The details of the texture are generated in a coarse-to-fine way; the initial scale generates the general color and macro-scale features and additional scales add on fine-grain details in an image. However, it is possible to create replica textures for larger values of $K$. Of the $34$ images generated for the experiments, there were four repetitions when $K = 2$ for $256 \times 256$ images. See Fig 4. below for an example.
\begin{figure}[hbt]
\begin{minipage}[t]{0.2375\columnwidth}
  \includegraphics[width=\linewidth]{refs/img_24.jpg}
\end{minipage}\hfill 
\begin{minipage}[t]{0.2375\columnwidth}
  \includegraphics[width=\linewidth]{k_eq_0/result_24.jpg}
\end{minipage}\hfill
\begin{minipage}[t]{0.2375\columnwidth}
  \includegraphics[width=\linewidth]{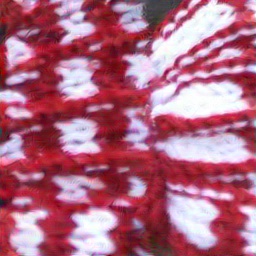}
\end{minipage}\hfill 
\begin{minipage}[t]{0.2375\columnwidth}
  \includegraphics[width=\linewidth]{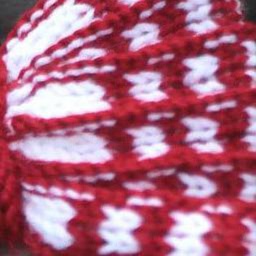}
\end{minipage}\hfill 
\caption{Progression of synthesis that lead to repetitions. \textbf{Left:} Reference Texture. \textbf{Middle Left:} $K = 0$. \textbf{Middle Right:} $K = 1$. \textbf{Right:} $K = 2$.}
\label{fig: mutliscale repetition} 
\end{figure} Based on results for generating large textures from \cite{gonthier}, we believe the number of scales one can use before generating repetitions depends on the resolution of the image. Our tests found no repetitions when $K = 1$ for $256 \times 256$ images.

\begin{figure}[hbt]
\begin{minipage}[t]{0.2375\columnwidth}
  \includegraphics[width=\linewidth]{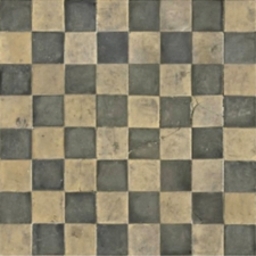}
\end{minipage}\hfill 
\begin{minipage}[t]{0.2375\columnwidth}
  \includegraphics[width=\linewidth]{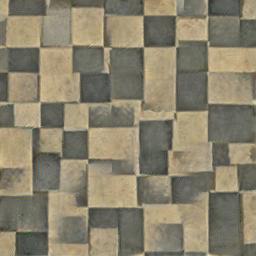}
\end{minipage}\hfill 
\begin{minipage}[t]{0.2375\columnwidth}
  \includegraphics[width=\linewidth]{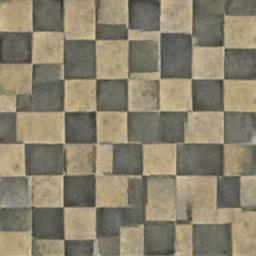}
\end{minipage}\hfill
\begin{minipage}[t]{0.2375\columnwidth}
  \includegraphics[width=\linewidth]{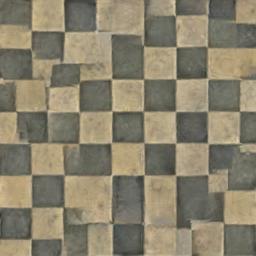}
\end{minipage}\hfill 
\begin{minipage}[t]{0.2375\columnwidth}
  \includegraphics[width=\linewidth]{refs/img_33.jpg}
\end{minipage}\hfill 
\begin{minipage}[t]{0.2375\columnwidth}
  \includegraphics[width=\linewidth]{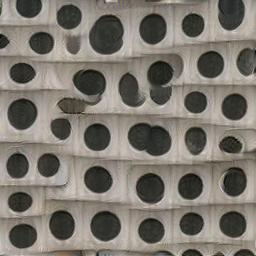}
\end{minipage}\hfill 
\begin{minipage}[t]{0.2375\columnwidth}
  \includegraphics[width=\linewidth]{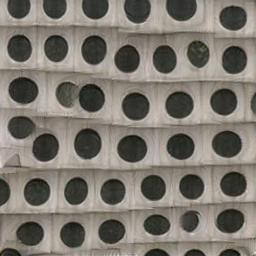}
\end{minipage}\hfill 
\begin{minipage}[t]{0.2375\columnwidth}
  \includegraphics[width=\linewidth]{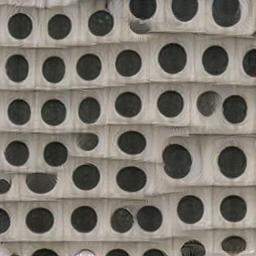}
\end{minipage}
\caption{Results with SW Loss. \textbf{Left:} Reference. \textbf{Mid Left:} $K = 0$. \textbf{Mid Right:} $K = 1$. \textbf{Right:} $K = 2$.}
\label{fig: no loss} 
\end{figure}

In the Fig. \ref{fig: no loss}, a small ablation study is done to test the effectiveness of \cref{eqn: new slicing loss}. The multi-scale approach is applied without the additional loss term in \cref{eqn: new slicing loss}. 
 From our results, the multi-scale approach by itself is not enough to fully capture nonstationary statistics or enforce long range constraints, and0 the loss term added in \cref{eqn: new slicing loss} is the factor accounting for improvements in synthesis quality.

\subsection{Additional comparisons}
Our proposed algorithm with $K = 1$ is compared with the synthesis using SW, a gram matrices with a spectrum constraint, our proposed method with $K = 0$, and the multi-scale method from \cite{gonthier}, which uses gram matrices with a spectrum constraint. In Fig. 6. we provide synthesis results, and we provide a quantitative study in Table 2.

\begin{table}[h]
\caption{Comparison of Various Synthesis Methods}
\centering
\begin{tabular}{||c c c c c c||} 
\hline
Method & LPIPS  & FID & c-FID & KID & c-KID \\ [0.25ex] 
 \hline
  $K = 0$  & $0.437$ & $107.220$ & $71.938$ & $-0.014$ & $0.073$\\ 
 \hline
 SW  & $0.454$ & $101.768$ & $78.683$ & $-0.016$ & $0.083$\\
 \hline
 Spec. & $0.447$ & $99.615$ & $78.250$ & $-0.016$& $0.083$\\
 \hline
 Gonthier  & $0.415$ & $77.569$ & $67.728$ & $\mathbf{-0.018}$ & $0.067$\\
 \hline
$K=1$ & $\mathbf{0.381}$ & $\mathbf{67.118}$ & $\mathbf{53.908}$ & $\mathbf{-0.018}$ & $\mathbf{0.044}$\\ 
 \hline\hline
  GT & $0$ & $0$ & $18.069$ & $-0.025$& $0$\\
  \hline
\end{tabular}
 \label{Tab: Final comparison}
\end{table}

\section{Conclusions}
We present a modification of texture synthesis via Sliced Wasserstein Loss, which yields a small set of statistics that has the ability to add long range constraints without user-added spatial tags or other forms of supervision. Our additional loss term can be thought of as a regularization term, but without the need for careful hyperparameter tuning. Future work would involve testing the dependence between number of scales and synthesis quality for a large dataset of textures.

\bibliographystyle{IEEEbib}

\newpage

\appendix

\onecolumn
\begin{figure}[h]
\begin{minipage}[t]{0.16\columnwidth}
  \includegraphics[width=\linewidth]{refs/img_30.jpg}
\end{minipage}\hfill 
\begin{minipage}[t]{0.16\columnwidth}
\includegraphics[width=\linewidth]{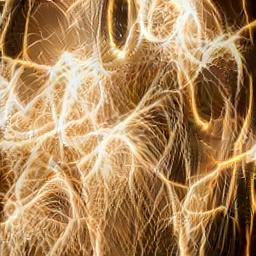}
\end{minipage}\hfill 
\begin{minipage}[t]{0.16\columnwidth}
\includegraphics[width=\linewidth]{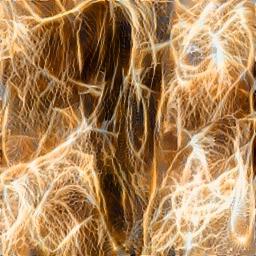}
\end{minipage}\hfill 
\begin{minipage}[t]{0.16\columnwidth}
\includegraphics[width=\linewidth]{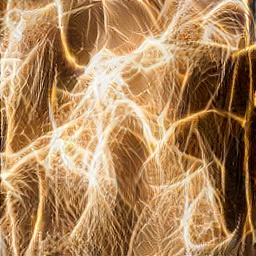}
\end{minipage}\hfill 
\begin{minipage}[t]{0.16\columnwidth}
\includegraphics[width=\linewidth]{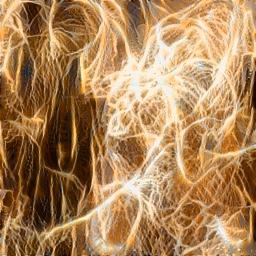}
\end{minipage}\hfill 
\begin{minipage}[t]{0.16\columnwidth}
  \includegraphics[width=\linewidth]{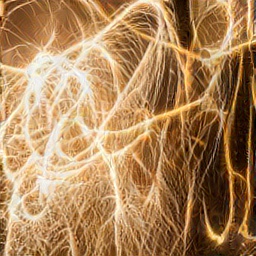}
\end{minipage}

\begin{minipage}[t]{0.16\columnwidth}
  \includegraphics[width=\linewidth]{refs/img_22.jpg}
\end{minipage}\hfill 
\begin{minipage}[t]{0.16\columnwidth}
\includegraphics[width=\linewidth]{heitz/result_22.jpg}
\end{minipage}\hfill 
\begin{minipage}[t]{0.16\columnwidth}
\includegraphics[width=\linewidth]{spectrum/result_22.jpg}
\end{minipage}\hfill 
\begin{minipage}[t]{0.16\columnwidth}
\includegraphics[width=\linewidth]{k_eq_0/result_22.jpg}
\end{minipage}\hfill 
\begin{minipage}[t]{0.16\columnwidth}
\includegraphics[width=\linewidth]{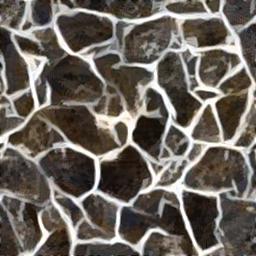}
\end{minipage}\hfill 
\begin{minipage}[t]{0.16\columnwidth}
  \includegraphics[width=\linewidth]{k_eq_1/result_22.jpg}
\end{minipage}

\begin{minipage}[t]{0.16\columnwidth}
  \includegraphics[width=\linewidth]{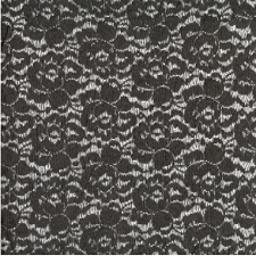}
\end{minipage}\hfill 
\begin{minipage}[t]{0.16\columnwidth}
\includegraphics[width=\linewidth]{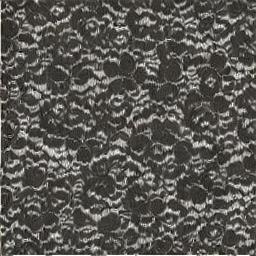}
\end{minipage}\hfill 
\begin{minipage}[t]{0.16\columnwidth}
\includegraphics[width=\linewidth]{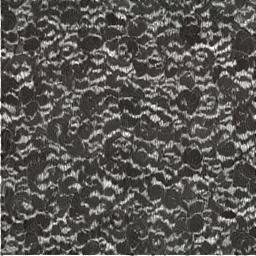}
\end{minipage}\hfill 
\begin{minipage}[t]{0.16\columnwidth}
\includegraphics[width=\linewidth]{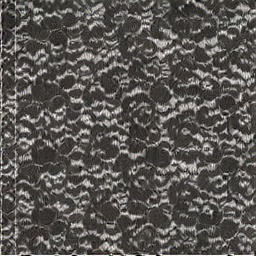}
\end{minipage}\hfill 
\begin{minipage}[t]{0.16\columnwidth}
\includegraphics[width=\linewidth]{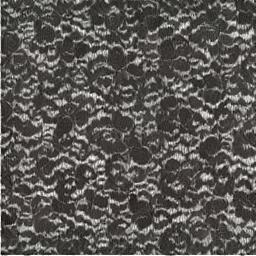}
\end{minipage}\hfill 
\begin{minipage}[t]{0.16\columnwidth}
  \includegraphics[width=\linewidth]{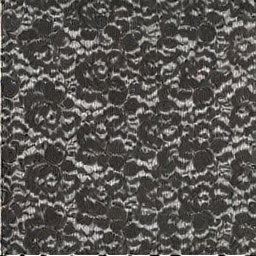}
\end{minipage}

\begin{minipage}[t]{0.16\columnwidth}
  \includegraphics[width=\linewidth]{refs/img_36.jpg}
\end{minipage}\hfill 
\begin{minipage}[t]{0.16\columnwidth}
\includegraphics[width=\linewidth]{heitz/result_36.jpg}
\end{minipage}\hfill 
\begin{minipage}[t]{0.16\columnwidth}
\includegraphics[width=\linewidth]{spectrum/result_36.jpg}
\end{minipage}\hfill 
\begin{minipage}[t]{0.16\columnwidth}
\includegraphics[width=\linewidth]{k_eq_0/result_36.jpg}
\end{minipage}\hfill 
\begin{minipage}[t]{0.16\columnwidth}
\includegraphics[width=\linewidth]{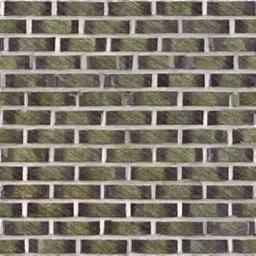}
\end{minipage}\hfill 
\begin{minipage}[t]{0.16\columnwidth}
  \includegraphics[width=\linewidth]{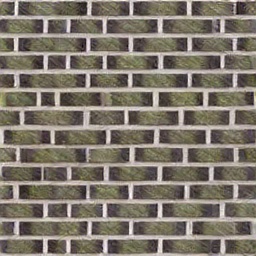}
\end{minipage}

\begin{minipage}[t]{0.16\columnwidth}
  \includegraphics[width=\linewidth]{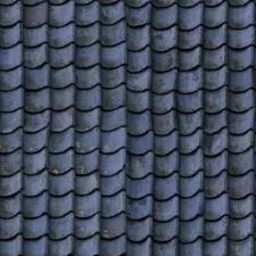}
\end{minipage}\hfill 
\begin{minipage}[t]{0.16\columnwidth}
\includegraphics[width=\linewidth]{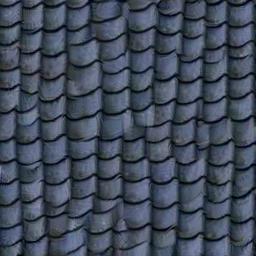}
\end{minipage}\hfill 
\begin{minipage}[t]{0.16\columnwidth}
\includegraphics[width=\linewidth]{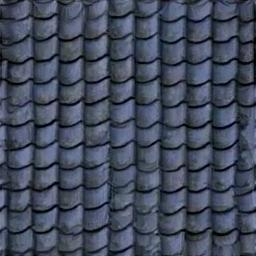}
\end{minipage}\hfill 
\begin{minipage}[t]{0.16\columnwidth}
\includegraphics[width=\linewidth]{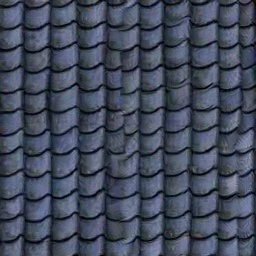}
\end{minipage}\hfill 
\begin{minipage}[t]{0.16\columnwidth}
\includegraphics[width=\linewidth]{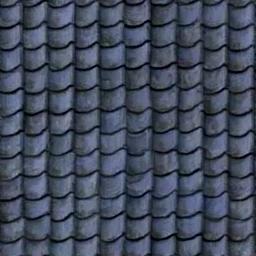}
\end{minipage}\hfill 
\begin{minipage}[t]{0.16\columnwidth}
  \includegraphics[width=\linewidth]{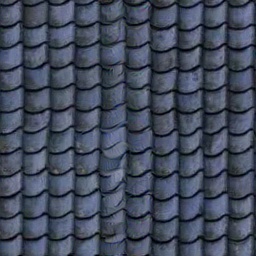}
\end{minipage}

\begin{minipage}[t]{0.16\columnwidth}
  \includegraphics[width=\linewidth]{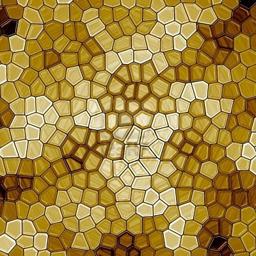}
\end{minipage}\hfill 
\begin{minipage}[t]{0.16\columnwidth}
\includegraphics[width=\linewidth]{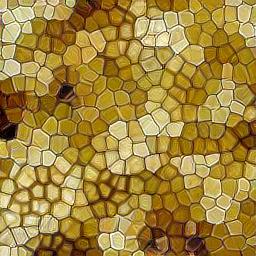}
\end{minipage}\hfill 
\begin{minipage}[t]{0.16\columnwidth}
\includegraphics[width=\linewidth]{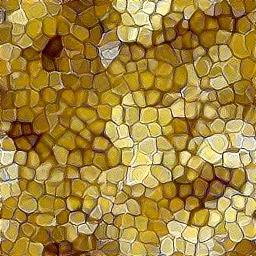}
\end{minipage}\hfill 
\begin{minipage}[t]{0.16\columnwidth}
\includegraphics[width=\linewidth]{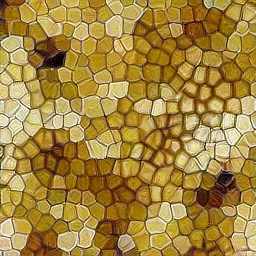}
\end{minipage}\hfill 
\begin{minipage}[t]{0.16\columnwidth}
\includegraphics[width=\linewidth]{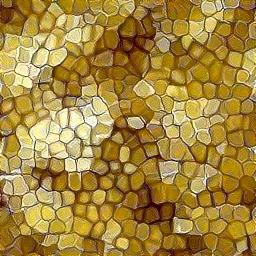}
\end{minipage}\hfill 
\begin{minipage}[t]{0.16\columnwidth}
  \includegraphics[width=\linewidth]{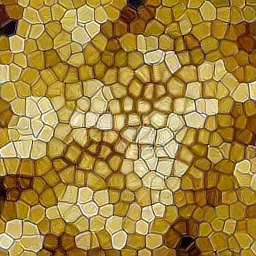}
\end{minipage}

\begin{minipage}[t]{0.16\columnwidth}
  \includegraphics[width=\linewidth]{refs/img_33.jpg}
\end{minipage}\hfill 
\begin{minipage}[t]{0.16\columnwidth}
\includegraphics[width=\linewidth]{heitz/result_33.jpg}
\end{minipage}\hfill 
\begin{minipage}[t]{0.16\columnwidth}
\includegraphics[width=\linewidth]{spectrum/result_33.jpg}
\end{minipage}\hfill 
\begin{minipage}[t]{0.16\columnwidth}
\includegraphics[width=\linewidth]{k_eq_0/result_33.jpg}
\end{minipage}\hfill 
\begin{minipage}[t]{0.16\columnwidth}
\includegraphics[width=\linewidth]{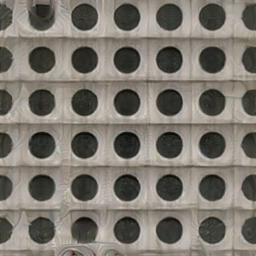}
\end{minipage}\hfill 
\begin{minipage}[t]{0.16\columnwidth}
  \includegraphics[width=\linewidth]{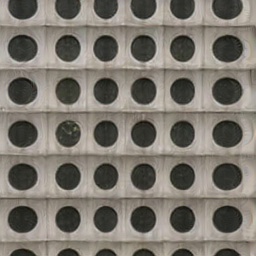}
\end{minipage}

\begin{minipage}[t]{0.16\columnwidth}
  \includegraphics[width=\linewidth]{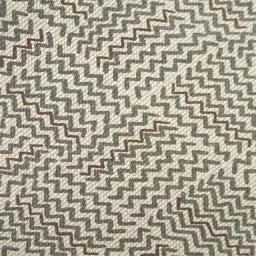}
\end{minipage}\hfill 
\begin{minipage}[t]{0.16\columnwidth}
\includegraphics[width=\linewidth]{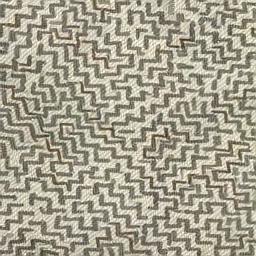}
\end{minipage}\hfill 
\begin{minipage}[t]{0.16\columnwidth}
\includegraphics[width=\linewidth]{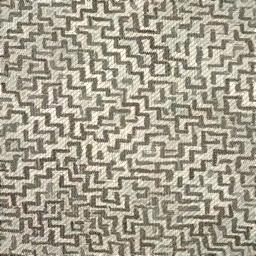}
\end{minipage}\hfill 
\begin{minipage}[t]{0.16\columnwidth}
\includegraphics[width=\linewidth]{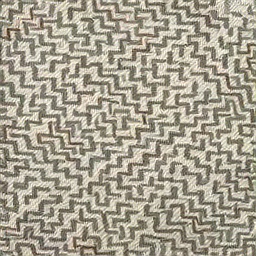}
\end{minipage}\hfill 
\begin{minipage}[t]{0.16\columnwidth}
\includegraphics[width=\linewidth]{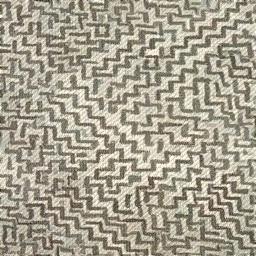}
\end{minipage}\hfill 
\begin{minipage}[t]{0.16\columnwidth}
  \includegraphics[width=\linewidth]{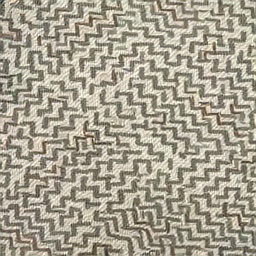}
\end{minipage}
\caption{Additional visual comparison of results. \textbf{First Column:} Reference. \textbf{Second Column:} SW Loss. \textbf{Third Column:} Spectrum Constraint. \textbf{Fourth Column:} $K = 0$ (Ours). \textbf{Fifth Column:} Gonthier. \textbf{Sixth Column:} $K = 1$ (Ours).}
\label{fig: more examples} 
\end{figure}
\end{document}